\documentclass{article}

\PassOptionsToPackage{square,numbers}{natbib}
\usepackage[final, dandb]{neurips_2025}

\usepackage[utf8]{inputenc} %
\usepackage[T1]{fontenc}    %
\usepackage{hyperref}       %
\usepackage{url}            %
\usepackage{booktabs}       %
\usepackage{amsfonts}       %
\usepackage{nicefrac}       %
\usepackage{microtype}      %
\usepackage{xcolor}         %
\usepackage{authblk}
\usepackage{enumitem}
\usepackage{amsmath}
\usepackage{graphicx}
\setlist[itemize]{topsep=0pt}  %
\setlist[enumerate]{topsep=0pt}
\setlist[itemize,enumerate]{leftmargin=*}

\usepackage{tabularray}
\usepackage{multirow}

\title{EmoSign: A Multimodal Dataset for Understanding Emotions in American Sign Language}

\author{%
Phoebe Chua$^{1, 2*}$ \quad Cathy Mengying Fang $^{1*}$ \quad Takehiko Ohkawa $^3$ \quad Raja Kushalnagar $^4$ \quad Suranga Nanayakkara$^2$ \quad Pattie Maes $^1$ \\
$^1$MIT Media Lab \quad $^2$National University of Singapore \quad $^3$The University of Tokyo \quad  $^4$Gallaudet University \\
\texttt{\{phochua, catfang, pattie\}@mit.media.edu}\\
\texttt{\{pchua, scn\}@nus.edu.sg}\\
\texttt{ohkawa-t@iis.u-tokyo.ac.jp}\\
\texttt{raja.kushalnagar@gallaudet.edu}
}

\begin{document}

\maketitle
\def\thefootnote{*}\footnotetext{These authors contributed equally to this work.}

\begin{abstract}
Unlike spoken languages where the use of prosodic features to convey emotion is well studied, indicators of emotion in sign language remain poorly understood, creating communication barriers in critical settings. Sign languages present unique challenges as facial expressions and hand movements simultaneously serve both grammatical and emotional functions. To address this gap, we introduce EmoSign, the first sign video dataset containing sentiment and emotion labels for 200 American Sign Language (ASL) videos. We also collect open-ended descriptions of emotion cues. Annotations were done by 3 Deaf ASL signers with professional interpretation experience. Alongside the annotations, we include baseline models for sentiment and emotion classification. This dataset not only addresses a critical gap in existing sign language research but also establishes a new benchmark for understanding model capabilities in multimodal emotion recognition for sign languages. The dataset is made available at \url{https://huggingface.co/datasets/catfang/emosign}.

\end{abstract}

\section{Introduction} \label{sec:intro}

The emotional content of a speech comes not only from its linguistic content but also \textit{how} it is spoken---from pitch and intonations to non-verbal expressions \cite{cutler1997prosody}. For sign language, however, emotional indicators are less understood and studied \cite{elliott2013facial,hietanen2004perception,lim2024exploring}.
This ambiguity has practical negative consequences from misinterpretations of signers' feelings to causing biases and prejudices in legal settings \cite{taira2019hearing} and emergency departments \cite{james2022they}.
The root of the ambiguity comes from the differences in the makeup of sign languages. Sign languages are comprised of five parameters: 
Handshape, Place of Articulation (where the sign is made), Movement (how the articulators move), Orientation (the hands’ relation towards the Place of Articulation), and Non-manual behaviors (what the body and face are doing). They function similarly to the cavities, articulators and features of spoken languages \cite{valli2000linguistics,pfau2012sign}. The non-manual behaviors serve as linguistic (e.g., grammatical) markers in addition to conveying emotions. One category of such non-manual behaviors is facial expressions. For example, ``puffing of one's face'' indicates intensity  and ``raised eyebrows'' indicates a yes/no question \cite{valli2000linguistics}. At the same time, one's facial expressions also convey emotions, which are applied on top of the linguistic meaning \cite{lim2024exploring}. Other ways of expression emotions in sign languages include the hand movements, such as the tempo, rhythm, and size of the movements \cite{reilly1992affective}. For example, the duration of sentences was shorter, the sign movements were more angular, and their movement paths shortened when expressing anger as compared to a neutral condition.

Emotion recognition is a common task in the machine learning community. Datasets and models have been constructed to recognize the sentiment and emotions based on text \cite{hutto2014vader,tan2022roberta}, speech \cite{ma2023emotion2vec,busso2008iemocap}, facial expressions \cite{baltruvsaitis2016openface}, and a combinations of multiple input modalities \cite{zadeh2018multimodal,chen2017multimodal,poria2018meld}. Emerging multi-modal foundation models have demonstrated abilities to understand context of the videos \cite{liu2023improved}. Few have extended the capabilities of pre-trained models to understand affect via fine-tuning and instruction-tuning \cite{li2024eald,cheng2024emotion} on new datasets and benchmarks \cite{lian2025affectgpt}. Emotion recognition on sign languages, however, remain sparse. Sign language translation itself is already difficult. In addition, current approaches fail to recognize the complex linguistic and affective functions of facial expressions during signing.

To address this critical research gap, we introduce EmoSign, the first comprehensive dataset containing sentiment and emotion labels for American Sign Language (ASL) videos. The dataset includes 200 ASL video clips annotated by 3 Deaf ASL signers with professional interpretation experience, who provided: (1) Overall sentiment ratings on a 7-point scale, (2) Presence and intensity ratings for 10 distinct emotion categories (3) Detailed descriptions of specific emotion cues. Unlike existing sign language datasets that focus primarily on translation capabilities, EmoSign specifically targets the affective dimensions of signing.
Our work contributes to both sign language accessibility and emotion recognition research by:

\begin{enumerate}
    \item Providing the first dedicated dataset for studying emotional expression in ASL
    \item Documenting descriptions of how emotions manifest through manual and non-manual components through the lens of native signers
    \item Establishing baseline model performance for sentiment analysis and emotion classification tasks in sign language
\end{enumerate}

The impact of our work extends beyond emotion recognition of sign languages. Studying how emotions can be recognized from a mix of non-manual and manual components allows us to study emotion differently. This also presents a new technical challenge to the AI/ML community with a challenging new benchmark for multimodal AI systems that tests their ability to distinguish between grammatical and affective functions of emotional cues in sign languages.

\section{Related Work} \label{sec:relwork}
\subsection{Emotion recognition} %
Datasets for emotion recognition are typically annotated using either discrete or dimensional emotion frameworks. Discrete emotion labels are based on theories of basic emotion, which suggest that there are a limited number of emotion states (e.g., fear, happiness)  associated with distinct expressions and physiological states \cite{ekman1992argument}. In contrast, dimensional approaches quantify emotion along continuous axes, predominantly emotional arousal and valence \cite{cowen2017self}. Recent advances in large language models (LLMs) have expanded the scope of emotion recognition beyond the traditional paradigm of emotion label prediction. These models facilitate a more generative approach to emotion \textit{understanding}, producing detailed, comprehensive descriptions of emotional states in natural language \cite{lian2025affectgpt}. This shift has prompted the development of new datasets and metrics that accommodate rich natural language descriptions of emotions, allowing for greater nuance in emotion analysis.

Many approaches to recognizing human emotions have used facial expressions \cite{lee2019context}, speech or text sentiment. Increasingly, multimodal approaches that combine video, audio, text and image inputs are being explored as a way of improving model robustness in complex environments \cite{lian2023mer}. However, many existing multimodal models are still heavily reliant on \cite{liang2024hemm}, or biased towards \cite{xiao2024can}, the language modality. To address this limitation, recent work has begun investigating novel sources of data that capture nonverbal social cues without relying on language, such as mime videos \cite{li2025mimeqa}. Sign language videos are also a rich source of expressive nonverbal data. They present unique challenges, but also opportunities, for multimodal emotion recognition. In hearing communities, many emotional expressions are universal in the sense that they can be reliably understood by people across many different cultures \cite{cordaro2014universals}. However, in sign languages, facial expressions and other non-manual components (such as mouth shapes and body language) often simultaneously serve grammatical and emotional functions. As such, recognizing the emotion of sign videos is often challenging for non-signers \cite{lim2024exploring}.

\subsection{Machine learning research on sign language}
In the context of machine learning research, work on sign language has focused on sign language translation and production. Sign language translation (SLT) methods typically use either raw image data or skeletal representation of the signer's pose as input. In terms of model architectures, Transformer-based architectures have been used for word-level sign language recognition based on 2D body pose sequence representations \cite{bohavcek2022sign}. The translation capabilities of LLMs also appear to extend to sign languages. SignLLM \cite{fang2024signllm} proposes a framework for transforming sign videos into language-like representations that can easily be passed to off-the-shelf LLMs. To do so, the sign videos are first processed into a sequences of discrete character-level sign tokens using a learned codebook. Then, the tokens are composed into word-level sign tokens to form a sentence. LLaVA-SLT \cite{liang2024llava} takes a slightly different approach. First, they pretrain an LLM using paired gloss-text data to learn the grammatical structure of sign language as well as its relationship with natural language. Next, they pretrain a visual encoder using about 400 hours of video-text pairs. Finally, a simple two-layer neural network is used to map the pretrained visual language embeddings into the LLM's token embedding space and produce a translation. In parallel with SLT, research has also investigated sign language production (SLP), which has useful applications such as automatic sign language captioning. SLP typically involves converting text to gloss (a method of sign language transcription), mapping the gloss to pose, then rendering the pose into a video or avatar \cite{fang2024signllm}. Despite significant advances in both SLT and SLP, challenges remain in developing systems that accurately capture and convey emotional nuance in sign language.

To the best of our knowledge, there does not exist a dataset that captures the emotion labels of sign languages. We hope to contribute to the expansion of current foundational models' capabilities and set up a new benchmark that have specific societal impacts.

\section{EmoSign: Multimodal Dataset of Sign Language with Emotion Labels} \label{sec:5}

\subsection{Dataset Collection and Pre-processing}

\begin{table}[]
\resizebox{\textwidth}{!}{%
\begin{tabular}{llllll}
\hline
Dataset       & Size              & Signers            & ASL Fluency          & Source & Labels            \\ \hline
YouTube-ASL \cite{uthus2023youtube}  & 984h              & \textgreater{}2500 & Mixed/unknown        & Web    & English captions  \\
OpenASL \cite{shi2022open}  & 288h              & 220                & DHH and interpreters & Web    & English captions  \\
ASL STEM Wiki \cite{yin2024asl} & \textgreater{}300h &  37                 & Interpreters         & Lab    & English sentences \\
ASLLRP \cite{neidle2022asl}  &
  \begin{tabular}[c]{@{}l@{}}2,651 \\ utterances\end{tabular} &
  19 &
  ASL native signers &
   &
  \begin{tabular}[c]{@{}l@{}}English and gloss captions, \\ non-manual information\end{tabular} \\
How2Sign \cite{duarte2021how2sign} & 79h               & 11                 & Interpreters         & Lab    & English captions  \\ 
MS-ASL \cite{joze2018ms} & 24h & 222 & Mixed/unknown & Web & English caption \\
\hline
\end{tabular}%
}
\caption{Overview of existing American Sign Language (ASL) datasets.}
\label{tab:asl-datasets}
\vspace{-18px}
\end{table}

Table \ref{tab:asl-datasets} summarizes existing continuous signing ASL datasets. We excluded YouTube-ASL due to the uncertainty of the quality of the signing and captions in the dataset. For the remaining datasets, we sampled several hours of video from each and segmented the videos into sentence or utterance-long clips. We then used VADER \cite{hutto2014vader} to calculate the text sentiment of each clip's caption. Across the board, we found that a large majority of videos were associated with neutral or close-to-neutral text captions. Based on this preliminary review, we selected ASLLRP as the base dataset as it provides the most comprehensive and high-quality labels for each video and also contains videos with strong emotional intensity. 

To address the limitations of existing ASL datasets, we collect additional emotion labels of sentiment, emotion and open-ended visual emotion cues using continuous signing videos from ASLLRP \cite{neidle2022asl} as the base. The annotation process is costly due to the manual effort involved in annotating multiple emotion labels for each video, as well as the difficulty of recruiting ASL-native signers for the task. To manage costs, we selected a sample of 200 utterances from ASLLRP for annotation.

We first preselected a subset of ASLLRP's continuous signing videos based on a manual inspection of the text content for emotional expressiveness. The videos were then segmented into utterances, which typically consist of a single sentence or phrase, based on the utterance start and end frames provided in the dataset. As before, we used VADER to calculate the text sentiment of each utterance. Despite the pre-selection for emotionally expressive videos, a large proportion of utterances were still associated with neutral or close-to-neutral text captions. To achieve a range of labels in the dataset, we selected 100 utterances with the most positive and most negative VADER scores for inclusion in the final sample.

\subsection{Annotation Construction} \label{sec:annotation-construction}
This study was reviewed and approved by our Institutional Review Board, protocol number NUS-IRB-2024-1081. We recruited 3 ASL native signers through a third-party vendor to annotate the selected video clips (for more details about annotator recruitment, see Appendix \ref{app:annotation-construction}) . Prior to beginning the annotations, the signers attended a training session with the researchers. The goal of the session was to walk through the annotation process, align on the expected responses for each annotation task, and clarify and questions they may have regarding the tasks. The annotation process took roughly x hours per individuals. The annotations were collected via Qualtrics (See Appendix \ref{interface} for details about the annotation interface). The interface was refined through pilot tests with individuals whose first language is ASL (excluded from the final team of annotators).

For each video, the signers completed three annotation tasks in the following order: (1) sentiment analysis, (2) emotion classification, and (3) free response description of attributions of emotions. For the general sentiment, the annotators were asked to determine the overall sentiment of the video from strongly negative to strongly positive on a linear scale from $-3$ to $+3$ \cite{zadeh2018multimodal}. For emotion classification, the annotators were asked to determine the level of presence of 10 emotions from ``not present at all'' to ``extremely present'' on a scale of 0 to 3. The set of emotions are ``joy'', ``excited'', ``surprise (positive)'', ``surprise (negative)'', ``worry'', ``sadness'', ``fear'', ``disgust'', ``frustration'', and  ``anger''. The labels build on Ekman's basic emotions, \cite{ekman1992argument} and the circumplex model of affect \cite{russell1980circumplex}, and was also informed by prior work on emotion datasets that expand on basic emotion categories for richer annotations \cite{liu2022mafw}.

After the first two tasks, the annotators were also asked to rate their level of confidence regarding their scores on a scale of 0-100 (0: not confident at all; 100-extremely confident). Finally, the annotators were asked to describe specific cues that led them to identify the emotions they chose. Example descriptions were given about the speed and scale of movement, head and body movement, facial expressions, and signs that were emphasized. These guided questions were derived based on prior literature and interview results \cite{chua2025perspectives} as well as our pilot tests.
Note that we allow the annotators to skip any videos that they did not wish to annotate because of content or the quality of the videos.

\subsection{Dataset Post-processing}

Each label of the clip was labeled by minimally 1, maximally 3 annotators, given certain clips were skipped. For each label, we used the ``majority vote'' approach to find the most popular rating. In the case of a tie, we used the annotators' self-reported confidence score and selected the label from the most confident annotation. Table \ref{tab:agreement-score} shows the Krippendorff's alpha scores used to measure the agreement between the annotators of each label. Overall, the average score is 0.593, within the emotion categories, positive emotion labels had higher inter-annotator agreement than negative emotion labels.

\begin{table}[]
\centering
\resizebox{.7\textwidth}{!}{%
\begin{tabular}{@{}lrlrlr@{}}
\toprule
\textbf{label} & \multicolumn{1}{l}{\textbf{alpha}} & \textbf{label} & \multicolumn{1}{l}{\textbf{alpha}} & \textbf{label}   & \multicolumn{1}{l}{\textbf{alpha}} \\ \midrule
Sentiment      & 0.738                              & surprise\_neg  & 0.119                              & disgust          & 0.166                              \\
joy            & 0.699                              & worry          & 0.555                              & frustration      & 0.330                              \\
excited        & 0.552                              & sadness        & 0.333                              & anger            & 0.370                              \\
surprise\_pos  & 0.381                              & fear           & 0.351                              & \textbf{average} & 0.593                              \\ \bottomrule
\end{tabular}
}
\caption{Krippendorff's Alpha scores of each label of the Dataset on a scale of -1 to 1, where 1 indicates unanimous agreement and -1 indicating systematic disagreement.}
\label{tab:agreement-score}
\vspace{-18px}
\end{table}

\subsection{Final Dataset Analysis}

\begin{figure}[t!]
    \centering
    \begin{minipage}[t]{.48\textwidth}
        \centering
        \includegraphics[width=1\linewidth]{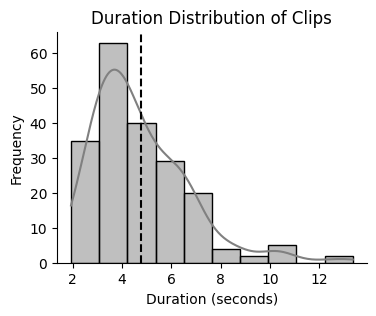}
        \caption{Duration distribution of the clips in the dataset. Dashed line indicates mean.}
        \label{fig:duration}
    \end{minipage}\hfill
    \begin{minipage}[t]{.48\textwidth}
        \centering
        \includegraphics[width=1\linewidth]{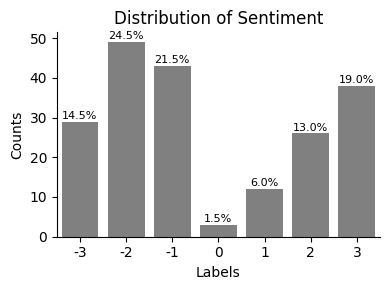}
        \caption{Distribution of sentiment labels. The labels correspond to the 7-point Likert scale where -3 is extremely negative, 0 is neutral, and 3 is extremely positive. Numbers above the bars indicate count.}
        \label{fig:sentiment-distrib}
    \end{minipage}%
    \vspace{-18px}
\end{figure}

The final dataset includes 200 utterances with an average length of 4.8s per utterance and a total of about 16 minutes of video. Figure \ref{fig:duration} shows the distribution of the duration of clips in the stimuli set. The dataset includes 4 different signers, and primarily depicts scenarios from everyday life such as conversations about the weather, family members and medical checkups.

\begin{figure} [b!]
\vspace{-18px}
    \centering
    \includegraphics[width=0.5\linewidth]{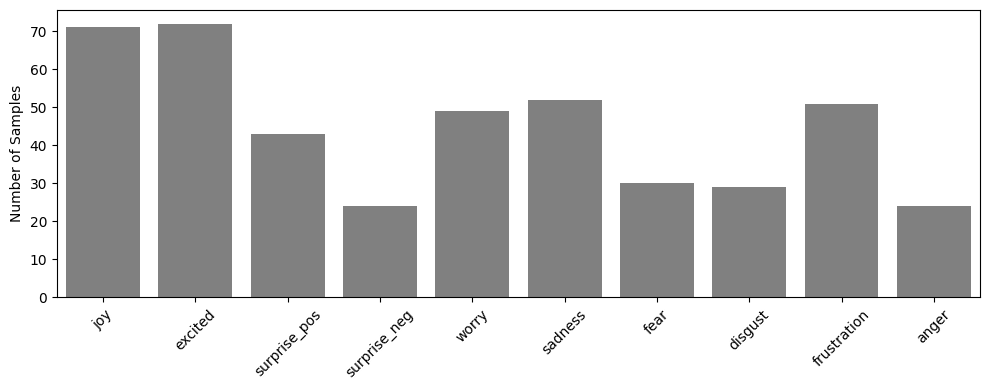}
    \caption{Distribution of emotion categories based on binarized presence across clips.}
    \label{fig:emotion-distribution}
\end{figure}

Figure \ref{fig:sentiment-distrib} shows the distribution of the sentiment labels. There are relatively few clips with neutral sentiment, but this is expected, since we selected clips with captions that had salient positive or negative emotions based on VADER. Figure \ref{fig:emotion-distribution} shows the distributions of the emotion categories. We binarized the presence of each emotion and the detailed breakdown distribution can be found in the Appendix \ref{app:data-details}. 

Looking at the annotators' response to attributions of emotional cues, we found some common themes which we elucidate in brief: (1) The non-manual markers are the primary cue for recognizing emotion in ASL. These includes \textit{facial expressions} such as furrowed brows, pursed lips, and squinted eyes, \textit{head movements} such as the head thrusts, tilting, and orientation changes that intensify the emotion, \textit{mouth movements} such as ``O'' shape, tongue out, and puffed lips, and \textit{body movement} such as shoulder raising and full-body tilting. (2) Signs were modified and emphasized for emotional expressions. Sign size (large/small), speed (fast/slow), repetition, and finger-spelling (sometimes for emphasis) are all noted as sign-based emotional markers. For stronger emotions (both positive and negative), signs are produced more broadly, quickly, or emphatically, very often in parallel with expressive non-manuals markers. (3) The role and context of the sentence is important to disambiguate emotions. Markers such as shifts in eye gaze, physical orientation, and changes in signing space are all used by signers to signal narrative switching or changes in perspective. The lack of context can cause uncertainty in emotion identification.

\section{Benchmarks}
\subsection{Benchmark Tasks}
We describe three benchmark tasks of increasing complexity that we carried out using the EmoSign dataset.

\textbf{Sentiment Analysis}. In this classification task, the goal is to predict the overall sentiment of a sign video on a 7-point Likert scale (-3 to 3), with a score of -3 corresponding to Strongly Negative and a score of +3 corresponding to Strongly Positive. In addition, we also formulated a coarse version of the task where the model is simply required to predict whether the sentiment is positive, neutral or negative. Following prior work \cite{zadeh2018multimodal,zadeh2017tensor}, we evaluate performance using accuracy (ACC) and weighted F1-score (WAF). We chose WAF as the primary metric and ACC as the secondary metric due to label imbalance.

\textbf{Single-label Emotion Classification}. Building on the sentiment analysis task, the goal of the emotion classification task is to predict the specific emotion that is most dominant or most present in a sign video. We first combined ``joy'' and ``excited'' into one category ``happiness'' due to their high co-occurrence within a single video (Jaccard similarity score of 0.81, Figure \ref{fig:jaccard} in Appendix). If all of the emotion categories were labeled as ``not present'', then a ``neutral'' label is assigned. We then separated the dataset into two sets for single-expression set and multi-expression set. The multi-expression set only contains classes that have more than 2 samples.  The single expression set consists of 140 clips and the multiple expression set consist of 37 clips. 
Appendix \ref{app:data-details} shows the distributions of the single and multiple expression set respectively. 

The model is required to select a single emotion label from a predefined set of ten possible labels (described in Section \ref{sec:annotation-construction}), as well as the label Neutral. Note that the model is not required to assess emotional intensity in this task. For each single-expression label, we present the accuracy and F1 score. Holistically, we chose weighted accuracy (wAcc) and weighted F1-score (wF1) as the evaluation metric based on prior approaches \cite{jiang2020dfew,liu2022mafw}.

\textbf{Emotion Cue Grounding}. The ability to accurately identify task-relevant temporal and spatial regions of a video, otherwise known as grounding, is a crucial component of video question-answering \cite{min2024morevqa} systems and has a wide range of valuable applications ranging from human-computer interaction \cite{wadley2022future} to clinical settings \cite{yang2012detecting}. Yet, there remains a significant performance gap between MLLMs and human annotators on visual grounding tasks \cite{xiao2024can}.

The goal of the emotion cue grounding task is to identify video frames and spatial regions relevant to the sentiment analysis and emotion classification tasks described above.

\subsection{Baselines} \label{sec:baselines}

We selected several multimodal LLMs (MLLMs) that support video-language inputs to obtain baseline results on the EmoSign dataset. Model cards and qualitative examples of model outputs are provided in Appendix \ref{app:models}. For all models and tasks, we conduct ablation studies to better understand the influence of individual modalities. Inference was conducted on a single 80GB NVIDIA Tesla A100 GPU.

We used GPT-4o to annotate the videos in the same way that a human annotator would have (See Appendix \ref{app:gpt-prompts} for prompts and sample outputs). Each video was sampled at 10 fps and coded in base64. We forced structured output where each API calls outputs responses to all three tasks. 

We also evaluated the performance of AffectGPT \cite{lian2025affectgpt}, Qwen2.5-VL-7B-Instruct \cite{bai2025qwen2} and MiniGPT4 \cite{ataallah2024minigpt4} on the benchmark tasks. AffectGPT is a multimodal LLM (MLLM) fine-tuned on various emotion recognition and understanding tasks. Qwen2.5-VL-7B-Instruct is a general-purpose MLLM optimized for instruction following, while MiniGPT4 is a vision-language model that uses an additional projection layer to align visual features with the language branch that displays strong performance on a range of benchmark tasks. In preliminary tests, we found that AffectGPT, Qwen2.5-VL-7B-Instruct and MiniGPT4 were unable to consistently produce clean output when prompted to respond to all three benchmark tasks at once. We adapted the prompts (detailed in Appendix \ref{app:mllm-prompts}) to improve the structure and interpretability of model outputs and conducted inference on each task separately.

\section{Results} \label{sec:results}
In this section, we present the baseline results of the selected MLLMs against the ground truth provided by the ASL native signers. One of the authors (hearing, not fluent in ASL) also annotated each video using the same interface, providing an estimate of how a human without access to the sign video captions would understand the emotions in the video. The experimental results revealed that providing access to the video captions almost always improved model performance; below, we provide detailed error analysis.

\subsection{Sentiment Analysis}
When only provided with the sign videos, models exhibited diverse biases and behaviors. AffectGPT consistently output sentiment as Neutral, suggesting an almost-complete lack of ability to recognize emotions in sign videos from visual cues alone. GPT-4o and Qwen2.5-VL-7B-Instruct tend to skew towards predicting positive sentiment, while we did not observe a consistent pattern for MiniGPT4. Interestingly, the hearing annotator demonstrated a tendency to perceive neutral clips as negative and positive clips as neural. These results align with previous study findings that hearing non-signers often misinterpret linguistic facial markers as indicators of negative emotion \cite{lim2024exploring}. 

When also given access to the video captions, GPT-4o shows improved performance but still retains a slight positive bias, and rarely selects the Neutral label. Both AffectGPT and Qwen2.5-VL-7B-Instruct have a tendency to predict simply Positive, Neutral or Negative even in the 7-class sentiment task, suggesting limitations in their ability to perform fine-grained sentiment analysis. MiniGPT4 shows a notable positive bias and appears to struggle to effectively integrate text data into its reasoning.

While hearing annotators tend to lean towards negative interpretations in the absence of text captions for additional context, the models, in general, exhibit a slight neutral-to-positive bias. This discrepancy highlights the gap in human versus machine perception in the domain of sign language sentiment analysis. A possible reason for this gap is that many foundational models are pre-trained with an emphasis on being helpful, harmless and honest \cite{bai2022training}, leading to a neutral or positive bias to mitigate potentially harmful or incorrect assertions about a person's emotion state, especially in ambiguous contexts. However, more research is required to fully understand these observed model behaviors.

\begin{table}[t!]
\centering
\resizebox{.7\textwidth}{!}{%
\begin{tabular}{@{}llrrrr@{}}
\toprule
\textbf{modality}                & \textbf{model}  & \multicolumn{2}{c}{\textbf{sentiment (3-class)}}   & \multicolumn{2}{c}{\textbf{sentiment (7-class)}}   \\ \midrule
                                 &                 & \multicolumn{1}{c}{wAcc} & \multicolumn{1}{c}{wF1} & \multicolumn{1}{c}{wAcc} & \multicolumn{1}{c}{wF1} \\ \midrule
\multirow{4}{*}{video}           & MiniGPT4        & 34.68                    & 40.00                   & 14.46                    & 13.03                   \\
                                 & Qwen 2.5        & 27.34                    & 16.47                   & 10.26                    & 2.44                    \\
                                 & AffectGPT       & 33.33                    & 0.04                    & 14.29                    & 0.04                    \\
                                 & GPT-4o          & 40.72                    & 24.43                   & 19.81                    & 5.97                    \\ \midrule
\multirow{4}{*}{video + caption} & MiniGPT4        & 21.65                    & 36.89                   & 9.76                     & 12.18                   \\
                                 & Qwen 2.5        & 41.10                    & 54.29                   & 15.84                    & 14.51                   \\
                                 & AffectGPT       & \textbf{56.18}           & 64.37                   & 21.02                    & 16.13                   \\
                                 & GPT-4o          & 52.13                    & \textbf{76.72}          & \textbf{22.89}           & \textbf{26.35}          \\\midrule
                                 & Hearing person* & 55.64                    & 57.64                   & 25.48                    & 21.39                   \\ \bottomrule
\end{tabular}

}
\caption{Benchmark results of sentiment analysis.}
\label{tab:sentiment-eval}
\vspace{-18px}
\end{table}

\subsection{Single-label Emotion Classification}
In the context of single-label emotion classification, when only provided with the sign videos, the models demonstrated limited ability to identify emotions beyond very broad and common categories. GPT-4o almost always classified videos as displaying either happiness or frustration, suggesting that it falls back to common emotional descriptors without the presence of text as a contextual guide. Similarly, AffectGPT limited its predictions mostly to happiness, sadness, or neutral emotions, Qwen2.5-VL-7B-Instruct to happiness and neutral, and MiniGPT4 predominantly classified videos as happy. 

Access to the video captions led to improved model performance, allowing for more accurate and nuanced emotional classification. Although GPT-4o still occasionally defaults to happiness and frustration, it shows enhanced capacity to distinguish emotions such as worry and disgust, and generally succeeded in identifying sentiment correctly. GPT-4o's tendency to favor labels such as worry and fear were aligned with the emotion co-occurrence patterns observed in the ground truth labels in EmoSign, suggesting that these emotions are relatively close together in the language embedding space \cite{huh2024platonic}.

AffectGPT still retained its tendency to give neutral predictions, though less so than before. It occasionally confused frustration for happiness, but otherwise generally succeeded in identifying overall sentiment correctly. Qwen2.5-VL-7B-Instruct showed improved performance on the emotion classification task, but developed a tendency to predict frustration. Like AffectGPT, it occasionally confused frustration for happiness. MiniGPT4 continued to display a bias towards labeling videos as happy, even for videos with a ground truth of negative emotions such as disgust. These persistent biases suggest the need for further model enhancements and fine-tuning on sign videos to improve their emotion recognition capabilities.

\begin{table}[t!]
\resizebox{\textwidth}{!}{%

\begin{tabular}{@{}llllllllllllll@{}}
\toprule
\textbf{modality}                & \textbf{model} & \textbf{HP} & \textbf{SP(P)} & \textbf{SP(N)} & \textbf{WR} & \textbf{SD} & \textbf{FR} & \textbf{DG} & \textbf{FS} & \textbf{AG} & \textbf{NE} & \textbf{total} & \textbf{total} \\ \midrule
                                 &                & Acc         & Acc            & Acc            & Acc         & Acc         & Acc         & Acc         & Acc         & Acc         & Acc         & wAcc           & wF1            \\ \midrule
\multirow{4}{*}{video}           & MiniGPT4       & 69          & 20             & 14             & 0           & 0           & 0           & 0           & 0           & 0           & 27          & 13.01          & 22.02          \\
                                 & Qwen 2.5       & 35          & 0              & \textbf{43}    & 0           & 0           & 0           & 0           & 5           & \textbf{33} & 27          & 14.39          & 18.53          \\
                                 & AffectGPT      & 11          & 0              & 0              & 7           & \textbf{30} & 0           & 0           & 5           & 0           & \textbf{73} & 12.62          & 11.03          \\
                                 & GPT-4o         & 35          & 0              & 0              & 7           & 0           & 0           & 20          & 53          & 0           & 0           & 11.50          & 20.76          \\ \midrule
\multirow{4}{*}{video + caption} & MiniGPT4       & \textbf{89} & 0              & 14             & 7           & \textbf{30} & \textbf{43} & 10          & 0           & \textbf{33} & 9           & 23.56          & 35.89          \\
                                 & Qwen 2.5       & 63          & \textbf{40}    & 29             & 64          & 0           & 14          & 20          & 32          & \textbf{33} & 55          & 34.96          & 44.67          \\
                                 & AffectGPT      & 85          & 20             & 0              & 50          & \textbf{30} & 14          & 10          & 32          & \textbf{33} & 27          & 30.17          & 47.77          \\
                                 & GPT-4o         & \textbf{89} & 20             & 0              & \textbf{79} & 20          & 29          & \textbf{50} & \textbf{74} & 0           & 0           & \textbf{35.97} & \textbf{55.09} \\ 
                                 \bottomrule
\end{tabular}

}
\caption{Benchmark results of single expression emotion classification. HP: happiness; SP(P): surprise (positive); SP(N): surprise (negative); WR: worry; SD: sadness; FR: fear; DG: disgust; FS: frustration; AG: anger; NE: neutral.}
\label{tab:emotion-eval}
\vspace{-18px}
\end{table}

\subsection{Reasoning analysis}
To obtain a preliminary understanding of model abilities to perform emotion cue grounding, we conducted a manual inspection of several randomly selected videos alongside the ground truth and each model's corresponding reasoning outputs.

Without captions, the inference outputs show that MiniGPT4 and GPT-4o are capable of identifying specific facial expressions within the sign videos and using them to reason about the signer's emotions, suggesting that these models can capture and interpret visual nuance to some extent. In contrast, AffectGPT and Qwen2.5-VL-7B-Instruct were only able to provide generic descriptions such as "The signer's facial expressions and body language do not convey strong emotions", or "The facial expressions are neutral, and the body language shows no particular direction or activity". 

With the captions provided, the models appear to use the linguistic context to guide their reasoning over the visual inputs. Despite a general improvement in task performance, we observed several failure modes: models sometimes misinterpreted the sentiment of the text caption, or correctly understood the text sentiment but evaluated the visual inputs in ways that diverged from the Deaf annotators. Furthermore, models would claim to identify specific cues, such as hand gestures and posture, that suggested certain emotions. When checking these cues against the sign videos, we observed that the cues recognized by GPT-4o, Qwen2.5-VL-7B-Instruct and MiniGPT4 (e.g., a thumbs-up sign) were truly present in the video. However, there was a recurring sense that the models were attempting to construct explanations that were consistent with their judgment of the text sentiment, rather than independently recognizing emotions from visual cues.

GPT-4o frequently repeated statements such as "relaxed body language" and "generally positive sentiment", indicating a possible over-reliance on common language patterns without truly consulting the visual context. Qwen2.5-VL-7B-Instruct often highlighted the lack of audio and exhibited reluctance to make definitive statements about the signer's emotion. Its reasoning sometimes demonstrated a lack of understanding about sign language as a concept, with outputs such as "the exact content of the sign language cannot be determined without audio." As in the sentiment and emotion classification tasks, AffectGPT's reasoning displays a bias towards the neutral label, frequently repeating statements such as "neutral expression" and "lack of obvious body language cues." Like Qwen2.5-VL-7B-Instruct, it was also hesitant to make statements about the signer's emotional state.

We observed a significant performance gap between the ground-truth labels and the MLLM predictions, especially when models are only given access to the visual modality. This gap underscores the difficulty of the benchmark tasks as well as the current limitations of MLLMs in comprehending the nonverbal emotion cues in sign language. These findings align with prior work indicating that MLLMs often struggle with visual understanding, and that strong performance on visual question-answering tasks are likely not due to genuine visual comprehension but rather a result of language shortcuts and spurious correlations with irrelevant visual information \cite{xiao2024can}.

\section{Limitations} \label{sec:limitations}
In this paper, we propose a new dataset of sign videos with sentiment, emotion and open-ended emotion cue labels, as well as benchmark tasks on the dataset.  We acknowledge several limitations of the present work. 

While VADER \cite{hutto2014vader} is an efficient way of performing sentiment analysis for large text datasets, its lack of sophistication in emotion understanding could have influenced the diversity and representativeness of selected videos. The sign videos around which we build EmoSign were derived from the ASLLRP continuous signing corpus \cite{neidle2022asl}, which does not include more complex contexts often present in real-world scenarios such as in-the-wild footage or videos featuring multiple speakers. Due to the high cost of manual annotations, we only selected 200 sign utterances for inclusion in the initial EmoSign dataset. As such, there are some restrictions in the range of emotions captured in the current dataset. Class imbalances pose an additional challenge due to the relatively small number of videos. To address many of these issues, we plan to expand the dataset size in future work. Further, we plan to expand the open-ended emotion cue annotations by incorporating spatial and temporal information, which could provide critical context for the emotion cue grounding task.

In terms of benchmarking, we evaluated a relatively small set of models, which may not capture the full range of capabilities and limitations across different model architectures and weights. We also did not evaluate model performance on multi-label emotion classification tasks \cite{liu2022mafw}. A broader evaluation across more models and tasks could provide a more comprehensive picture of current models' abilities to recognize emotion in sign language videos. Importantly, none of the models we tested were specifically trained on American Sign Language (ASL). Future work could investigate fine-tuning models on ASL data to improve recognition of sign gestures, which could directly support more robust emotion recognition.

\section{Conclusion} \label{sec:conclusion}
In this paper, we introduced EmoSign, the first multimodal dataset containing sentiment and emotion labels specifically for ASL videos. By providing annotations from Deaf native ASL signers on sentiment, emotion categories, and detailed descriptions of emotion cues, EmoSign addresses a critical gap in both sign language research and emotion recognition. Our benchmark evaluations of several state-of-the-art multimodal LLMs revealed significant limitations in their ability to recognize emotions in sign language videos, particularly when relying solely on visual input without text captions. We hope that EmoSign will encourage more research on the nuanced emotional expressions unique to sign languages as well as innovations in how multimodal models' ability to understand emotion-laden signed videos, leading to the creation of more emotionally expressive sign language interpretation.

\begin{ack}
We would like to acknowledge Paul Liang for providing us with feedback and support, and students from Gallaudet University who assisted with pilot tests of the annotation interface.

\end{ack}

\bibliographystyle{plainnat}
\bibliography{reference}

\newpage

\section*{Neurips Paper Checklist}

\begin{enumerate}

\item {\bf Claims}
    \item[] Question: Do the main claims made in the abstract and introduction accurately reflect the paper's contributions and scope?
    \item[] Answer: \answerYes{} %
    \item[] Justification: See Section \ref{sec:intro}.

\item {\bf Limitations}
    \item[] Question: Does the paper discuss the limitations of the work performed by the authors?
    \item[] Answer: \answerYes{} %
    \item[] Justification: See Section \ref{sec:limitations}.

\item {\bf Theory assumptions and proofs}
    \item[] Question: For each theoretical result, does the paper provide the full set of assumptions and a complete (and correct) proof?
    \item[] Answer: \answerNA{} %

\item {\bf Experimental result reproducibility}
    \item[] Question: Does the paper fully disclose all the information needed to reproduce the main experimental results of the paper to the extent that it affects the main claims and/or conclusions of the paper (regardless of whether the code and data are provided or not)?
    \item[] Answer: \answerYes{} %
    \item[] Justification: Yes, we provide the model cards in Appendix \ref{app:models} as well as the prompts used to generate the main experimental results in Appendix \ref{app:gpt-prompts} and \ref{app:mllm-prompts}.

\item {\bf Open access to data and code}
    \item[] Question: Does the paper provide open access to the data and code, with sufficient instructions to faithfully reproduce the main experimental results, as described in supplemental material?
    \item[] Answer: \answerYes{} %
    \item[] Justification: Yes, we provide access to the EmoSign dataset and code used for data processing. We also provide all prompts used to generate the main experimental results in Appendix \ref{app:gpt-prompts} and \ref{app:mllm-prompts}.

\item {\bf Experimental setting/details}
    \item[] Question: Does the paper specify all the training and test details (e.g., data splits, hyperparameters, how they were chosen, type of optimizer, etc.) necessary to understand the results?
    \item[] Answer: \answerNA{} %
    \item[] Justification: We did not run any experiments fine-tuning or training new models. 

\item {\bf Experiment statistical significance}
    \item[] Question: Does the paper report error bars suitably and correctly defined or other appropriate information about the statistical significance of the experiments?
    \item[] Answer: \answerNA{} %
    \item[] Justification: Our baseline evaluations did not include factors that require error bars, confidence intervals or statistical significance tests. 

\item {\bf Experiments compute resources}
    \item[] Question: For each experiment, does the paper provide sufficient information on the computer resources (type of compute workers, memory, time of execution) needed to reproduce the experiments?
    \item[] Answer: \answerYes{} %
    \item[] Justification: Section \ref{sec:baselines} specifies that all inference was conducted on a single 80GB NVIDIA Tesla A100 GPU.
    
\item {\bf Code of ethics}
    \item[] Question: Does the research conducted in the paper conform, in every respect, with the NeurIPS Code of Ethics \url{https://neurips.cc/public/EthicsGuidelines}?
    \item[] Answer: \answerYes{} %
    \item[] Justification: All annotators were paid fair wages to the best of our knowledge. The research was carried out after procedures were reviewed and approved by our IRB. As a sign language dataset, the videos inevitably contain personally identifiable information. We only release the video and utterance IDs of the clips included in the EmoSign dataset, following the access rules on the ASLLRP dataset which it is based on.

\item {\bf Broader impacts}
    \item[] Question: Does the paper discuss both potential positive societal impacts and negative societal impacts of the work performed?
    \item[] Answer: \answerYes{} %
    \item[] Justification: We discuss potential positive and negative societal impacts in Sections \ref{sec:intro}, \ref{sec:relwork} and \ref{sec:conclusion}.
    
\item {\bf Safeguards}
    \item[] Question: Does the paper describe safeguards that have been put in place for responsible release of data or models that have a high risk for misuse (e.g., pretrained language models, image generators, or scraped datasets)?
    \item[] Answer: \answerNA{} %
    \item[] Justification: We release an extension of a subset of the ASLLRP dataset \cite{neidle2022asl}. The utterances were created by ASL consultants so that they 1) cover a range of linguistic constructions (e.g., questions and conditional sentences), and 2) are natural for signers to produce \cite{neidle2018new}. There are no unsafe images or videos in the sign data, which are publicly available through a data access interface \footnote{\url{https://www.bu.edu/asllrp/indexright.html}}.

\item {\bf Licenses for existing assets}
    \item[] Question: Are the creators or original owners of assets (e.g., code, data, models), used in the paper, properly credited and are the license and terms of use explicitly mentioned and properly respected?
    \item[] Answer: \answerYes{} %
    \item[] Justification: The sign videos were derived from ASLLRP \cite{neidle2018new, neidle2022asl}. The license, terms of use and url of the source data is provided at our huggingface link: \url{https://huggingface.co/datasets/catfang/emosign}.

\item {\bf New assets}
    \item[] Question: Are new assets introduced in the paper well documented and is the documentation provided alongside the assets?
    \item[] Answer: \answerYes{} %
    \item[] Justification: Documentation is provided in Section \ref{sec:annotation-construction} and on Huggingface.

\item {\bf Crowdsourcing and research with human subjects}
    \item[] Question: For crowdsourcing experiments and research with human subjects, does the paper include the full text of instructions given to participants and screenshots, if applicable, as well as details about compensation (if any)? 
    \item[] Answer: \answerYes{} %
    \item[] Justification: The full text of instructions given to participants and screenshots of the annotation interface are available in Appendix \ref{app:annotation-construction}. We do not have the details of how much each individual annotator was paid for their work, as annotators were hired through a third-party vendor. However, we disclose the amount paid to the third-party vendor also in Appendix \ref{app:annotation-construction}

\item {\bf Institutional review board (IRB) approvals or equivalent for research with human subjects}
    \item[] Question: Does the paper describe potential risks incurred by study participants, whether such risks were disclosed to the subjects, and whether Institutional Review Board (IRB) approvals (or an equivalent approval/review based on the requirements of your country or institution) were obtained?
    \item[] Answer: \answerYes{} %
    \item[] Justification: We recruited human annotators to label sign videos for sentiment, emotion and free-response descriptions of emotion attributions. No risks outside of those normally encountered in everyday life were anticipated. The paper describes our IRB approval in \ref{sec:annotation-construction}.

\item {\bf Declaration of LLM usage}
    \item[] Question: Does the paper describe the usage of LLMs if it is an important, original, or non-standard component of the core methods in this research? Note that if the LLM is used only for writing, editing, or formatting purposes and does not impact the core methodology, scientific rigorousness, or originality of the research, declaration is not required.
    \item[] Answer: \answerYes{} %
    \item[] Justification: We describe our use of LLMs for benchmarking in Sections \ref{sec:baselines} and \ref{sec:results} as well as in Appendix \ref{app:gpt-prompts} and \ref{app:mllm-prompts}.

\end{enumerate}

\newpage
\appendix

\section{Appendix / supplemental material}

\subsection{Data Collection Additional Details}  \label{app:annotation-construction}
\begin{figure}[b!]
    \centering
    \begin{minipage}[t]{1\textwidth}
        \centering
        \includegraphics[width=1\linewidth]{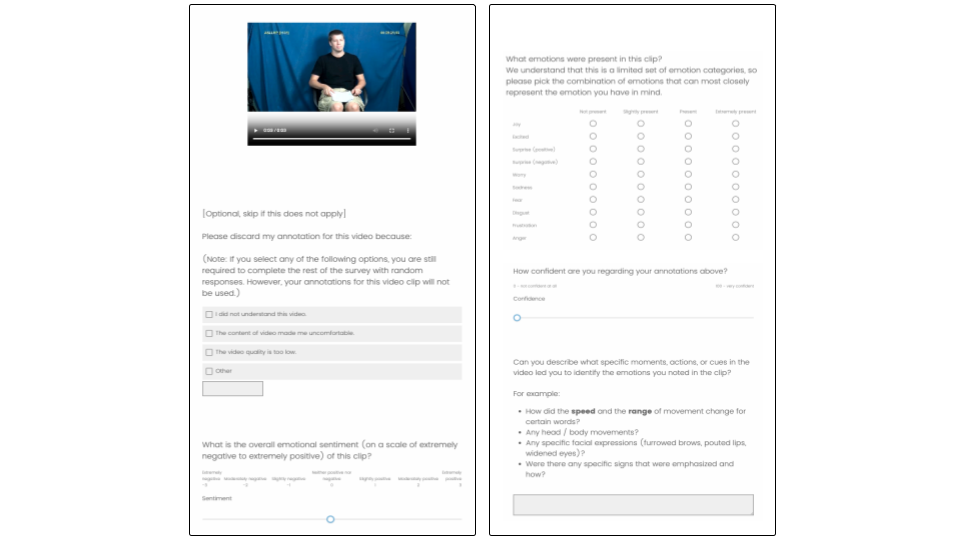}
        \caption{Annotation Interface}
        \label{fig:annotation-interface}
    \end{minipage}\hfill
    \begin{minipage}[t]{.39\textwidth}
        \centering
        \includegraphics[width=1\linewidth]{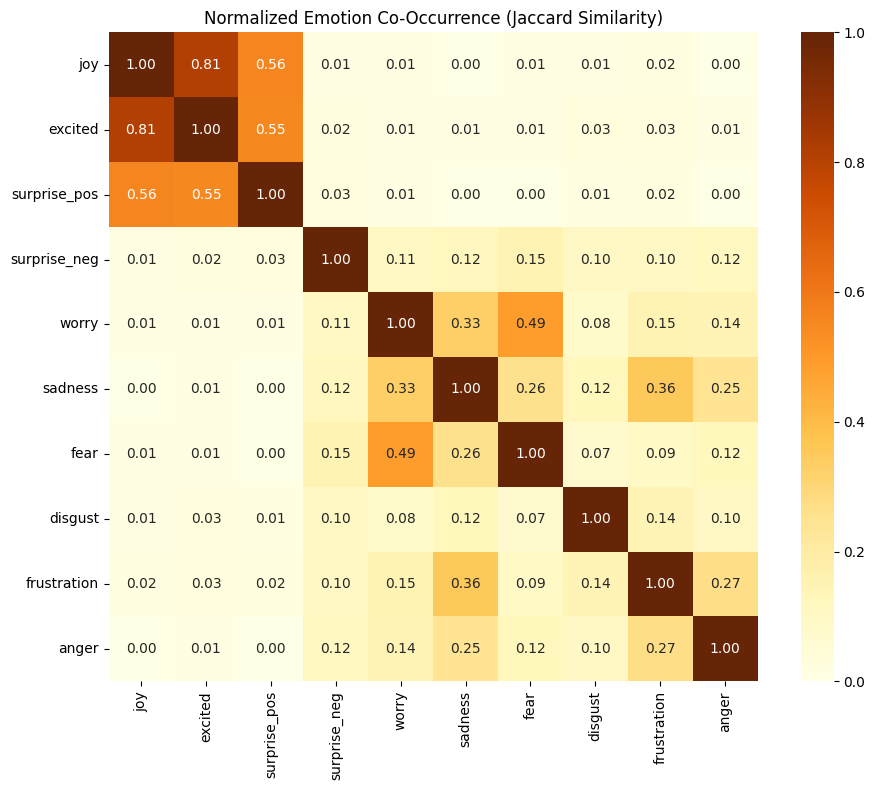}
        \caption{Jaccard Similarity of the original set of Emotion Labels.}
        \label{fig:jaccard}
    \end{minipage}%
\end{figure}

We employed a total of 3 annotators through a third-party vendor, specifically a full-service sign language interpreting and captioning company specializing in American Sign Language (ASL). The annotators utilized the interface shown in Figure \ref{fig:annotation-interface} for the annotation tasks. All three annotators completed all annotation tasks for all sign videos in the dataset. The vendor was paid \$400 per annotator.

\label{sec:interface}

\newpage

\subsection{Model Cards and Outputs} \label{app:models}

\begin{table}[h]
\resizebox{\textwidth}{!}{%
\begin{tabular}{@{}ll@{}}
\toprule
Model     & Link                                    \\ \midrule
Qwen2.5-VL-7B-Instruct \cite{bai2025qwen2} & \url{https://huggingface.co/Qwen/Qwen2.5-VL-7B-Instruct/tree/main} \\
Qwen3-8B \footnote{\url{https://github.com/QwenLM/Qwen3/blob/main/Qwen3_Technical_Report.pdf}}  & \url{https://huggingface.co/Qwen/Qwen3-8B}   \\
AffectGPT \cite{lian2025affectgpt} & \url{https://github.com/zeroQiaoba/AffectGPT} \\
MiniGPT4-video \cite{ataallah2024minigpt4} & \url{https://github.com/Vision-CAIR/MiniGPT4-video}                \\ \bottomrule
\end{tabular}%
}
\caption{Model cards for MLLMs used in EmoSign baseline results.}
\label{tab:modelcards}
\end{table}

\subsection{GPT-4o Details} \label{app:gpt-prompts}
The prompt template used for GPT-4o is as follows:

\noindent\fbox{%
    \parbox{\textwidth}{%
        \small
        You are an expert in the field of emotions. \\

        Please focus on facial expressions, body language, environmental cues, and events in the video and predict the emotional state of the character. Please ignore the character’s identity. We uniformly sample 10 frames per second from this clip. Please consider the temporal relationship between these frames. \\

        The video involves a person signing a sentence in ASL.  You have three tasks: \\           

        Task 1 - On a scale of extremely negative (-3) to extremely positive (+3), what is the overall affective sentiment of this clip? Output your classification as a number between -3 and 3.\\

        Task 2 - Which of the following affective categories were present in this clip? You may choose multiple options. The scale is from 0 to 3, where 0 is not present and 3 is extremely present. The possible categories are: joy, excited, surprise (positive), surprise (negative), worry, sadness, fear, disgust, frustration, anger. Output your classification as a json with the name of the emotion and a number between 0 and 3. \\

        Task 3 - Describe what specific moments, actions, or cues in the video that led you to your responses above? (e.g., the speed at which the person signed, specific facial expressions, content of the sign, etc). Output your 2-sentence response as a string.\\

        You should provide a structured response in the form of a Json string.
    }%
}

A sample response is formatted as follows: \\
\noindent\fbox{%
    \parbox{\textwidth}{%
        \small
    "filename": "00ADU7t7IWI\_1",\\
    "Sentiment": "2",\\
    "joy": "3", \\
    "excited": "2",\\
    ...\\
    "anger": "0", \\
    "QA": "The person is smiling and signing with energetic movements, indicating a positive and engaging message. The consistent smile and lively hand gestures suggest joy and excitement."
    }%
}

\newpage

\subsection{Other MLLM prompts for Sentiment and Emotion Recognition} \label{app:mllm-prompts}
The remaining MLLMs used for baseline evaluation struggled to follow the prompt template used for GPT-4o, which is relatively long and complex. As such, we created separate prompts for the sentiment analysis and single-label emotion recognition tasks presented in Section \ref{sec:results}.. Several of the multimodal LLMs used in the analysis had a dedicated method for incorporating video captions, in which case we used it to pass the caption to the mode. Otherwise, we included the caption directly within the prompt.

\noindent\fbox{%
    \parbox{\textwidth}{%
        \small
    {\textbf{Sentiment Prompt:} You are an expert in the field of emotions. The video involves a person signing a sentence in ASL. Please focus only on facial expressions and body language of the signer in the video and try to recognize the emotional state of the signer. Your task is: on a 7 point Likert scale ranging from extremely negative to extremely positive, what is the overall affective sentiment of this clip? You must select an answer from this list: [Extremely Negative, Negative, Somewhat Negative, Neutral, Somewhat Positive, Positive, Extremely Positive].} \\

    \textbf{Single-label Emotion Recognition Prompt:} You are an expert in emotion analysis. Your task is to analyze the given video clip and select the SINGLE MOST DOMINANT emotion from the following list:  [joy, excited, surprise (positive), surprise (negative), worry, sadness, fear, disgust, frustration, anger]. \\

    **Rules you MUST follow:**  
    
    1. **You must choose ONLY ONE emotion**—the one that is most prominent in the clip.  
    
    2. If multiple emotions seem present, select the strongest one.
    
    3. If unsure, make your best judgment—do not leave the response blank. \\

    **Penalties for not following instructions:**
    
    - If you output more than one emotion, you will fail the task.
    
    - If you refuse to answer or leave it blank, you will fail the task. \\

    **Output format:**  [selected emotion]  \\

    Now, analyze the clip and provide the SINGLE emotion label: \\

    \textbf{Sentiment extraction prompt:} Based on this text, what is the sentiment of the signer? The text is: [inference output]. Keep your answer brief without any explanation. \\

    \textbf{Emotion label extraction prompt:} Based on this text, what are the signer's emotions? The text is: [inference output]. Keep your answer brief without any explanation.

}}

\begin{figure}[h]
    \centering
    \includegraphics[width=\linewidth]{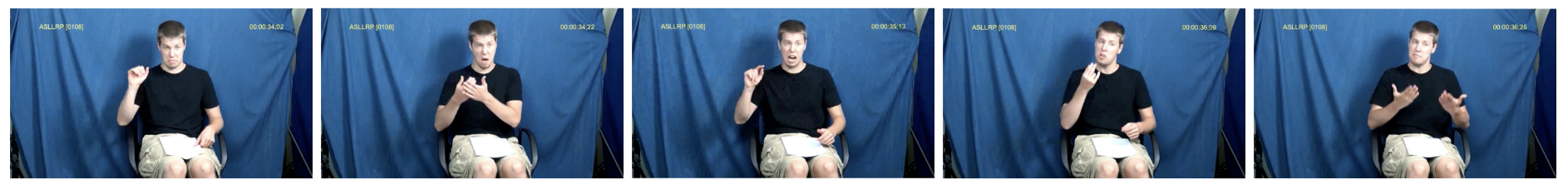}
    \caption{Sample frames from a sign video in the EmoSign dataset. The caption is: "If Mary gets home late, John will probably be upset."}
    \label{fig:cory-example}
\end{figure}

Figure \ref{fig:cory-example} provides an example of a sign video and the associated MLLM outputs from our baseline experiments.

\newpage

\fbox{\parbox{\dimexpr\textwidth-2\fboxsep-2\fboxrule\relax}{ \footnotesize

\textbf{Qwen2.5-VL-7B-Instruct (with captions):} Based on the signer's hand gestures and body language, it appears that they are expressing a sense of concern or worry about the situation described in the video caption. The signer's facial expression seems to convey a sense of anticipation or anxiety, which aligns with the idea of being upset if someone arrives late. Therefore, the most appropriate emotional state for the signer in this context would be "worry".\\

\textbf{Qwen2.5-VL-7B-Instruct (without captions):} Based on the signer's hand gestures and body language, it appears that they are expressing a sense of frustration or anger. The signer seems to be making emphatic movements with their hands, which could indicate a strong emotional response. However, without audio context, it's challenging to determine the exact nature of the emotion.\\

\textbf{AffectGPT (with captions):} The visual cues from the facial expressions and body movements of the signer, along with the caption "If Mary comes back late, John must be upset," suggest that the signer might be conveying a sense of worry or anxiety. This emotional state could be related to the potential impact of Mary's lateness on John. Therefore, the primary emotions displayed by the signer in the video might include worry, sadness, or anxiety. \\

\textbf{AffectGPT (without captions):} Given the video description, the signer's facial expressions and body language do not reveal any obvious signs of strong emotion. The head is slightly tilted, possibly indicating interest or concentration in the signing action. The eyes are closed, possibly to focus or concentrate on the signing, and the mouth is open, possibly due to speech or to express emotion. Considering these clues, it can be inferred that the signer might be in a calm or neutral emotional state. However, without additional information about the context of the signing or the individual's background, it is not possible to accurately determine the character's emotional state. \\

\textbf{MiniGPT4-video (with captions):} Mostly, John appears to be upset or frustrated due to Mary delaying returning home. His eyebrows are furrowed together, his lips curled downwards (shown by the wrinkles above them), as well as a slight clench of this jaw muscles which indicates stress or irritation - all these signs suggest he feels either anger/frustration (disgust) at her lateness. Secondarily one can see worry or anxiety expressed through brow wiggling while watching oneself being signing - there is an urge hidden behind those raised upper lids that could turn into full blown concern if things go wrong further... \\

\textbf{MiniGPT4-video (without captions):} The signer often exhibits a joyful expression during certain moments in the signature process. For example, when they finish signing an important sentence or document with satisfaction. They also show signs of happiness such as smiling wide and laughing occasionally which indicates positive emotions. There were instances wherein excitement is palpable from their energetic gestures particularly when discussing thrilling topics like travel experiences abroad or new technologies that arouse enthusiasm. \\

}}

\newpage
\subsection{Additional Descriptive Statistics of the Dataset and Benchmark Results} \label{app:data-details}

\begin{figure}[h]
    \centering
    \includegraphics[width=.7\linewidth]{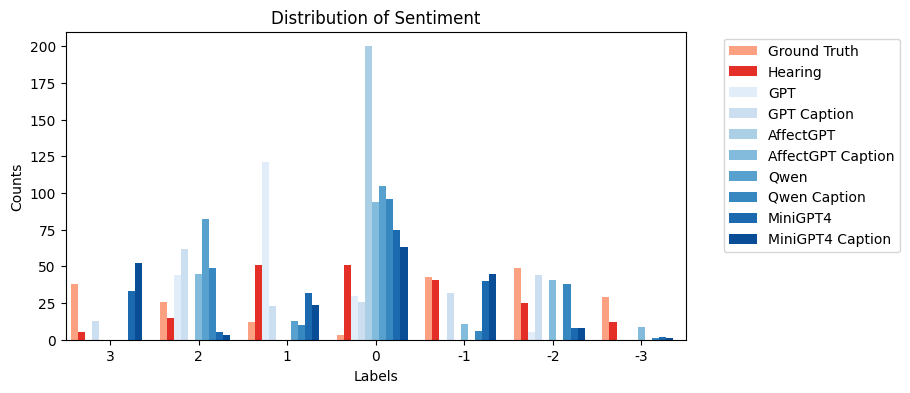}
    \caption{Distribution of Sentiment Analysis Results (7-class).}
    \label{fig:sent-dist-7}
\end{figure}

\begin{figure}[h]
    \centering
    \includegraphics[width=.7\linewidth]{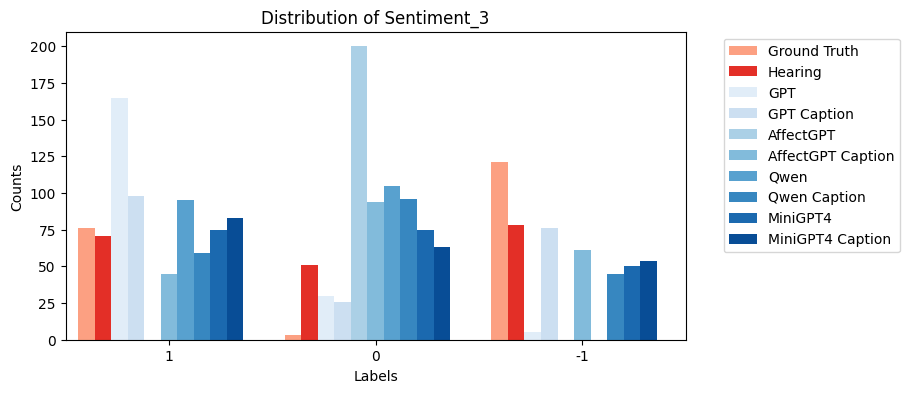}
    \caption{Distribution of Sentiment Analysis Results (3-class).}
    \label{fig:sent-dist-3}
\end{figure}

\begin{figure}[h]
    \centering
    \begin{minipage}[t]{.45\textwidth}
        \centering
        \includegraphics[width=1\linewidth]{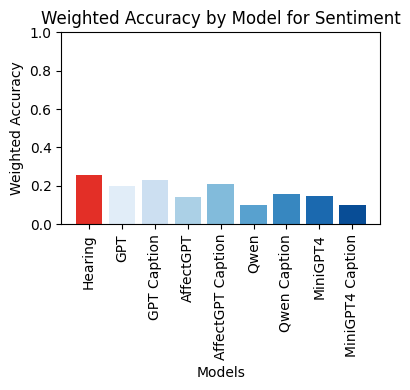}
        \caption{Weighted Accuracy of Models on Sentiment Analysis Task (7-class).}
        \label{fig:sent-wacc-7}
    \end{minipage}\hfill
    \begin{minipage}[t]{.45\textwidth}
        \centering
        \includegraphics[width=1\linewidth]{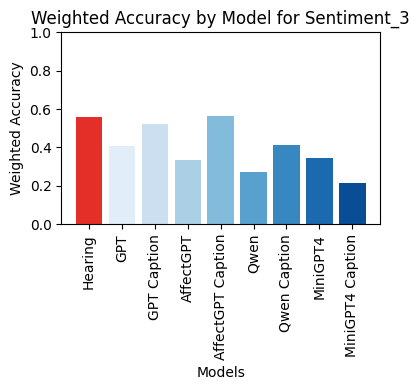}
        \caption{Weighted Accuracy of Models on Sentiment Analysis Task (3-class).}
        \label{fig:sent-wacc-3}
    \end{minipage}%

\end{figure}

\newpage

\begin{figure}[h]
    \centering
    \begin{minipage}[t]{.45\textwidth}
        \centering
        \includegraphics[width=1\linewidth]{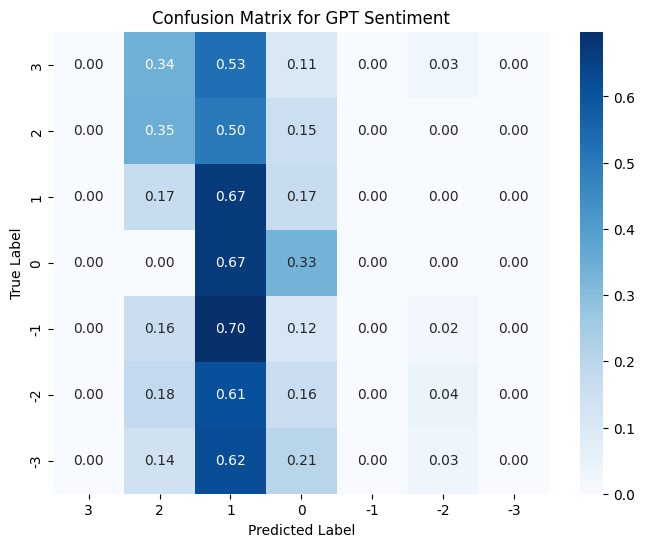}
        \caption{Confusion Matrix of GPT (without caption) output on Sentiment Analysis Task (7-class).}
        \label{fig:conf-sent-gpt}
    \end{minipage}\hfill
    \begin{minipage}[t]{.45\textwidth}
        \centering
        \includegraphics[width=1\linewidth]{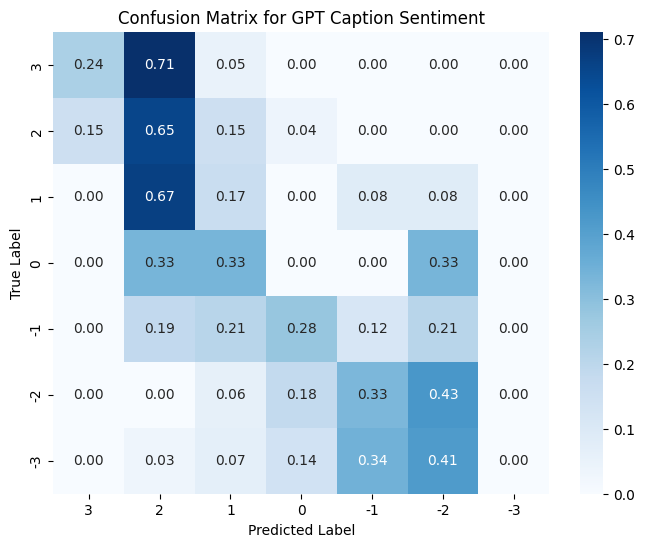}
        \caption{Confusion Matrix of GPT with caption output on Sentiment Analysis Task (7-class).}
        \label{fig:conf-sent-gpt-cap}
    \end{minipage}%

\end{figure}

\begin{figure}[h]
    \centering
    \begin{minipage}[t]{.45\textwidth}
        \centering
        \includegraphics[width=1\linewidth]{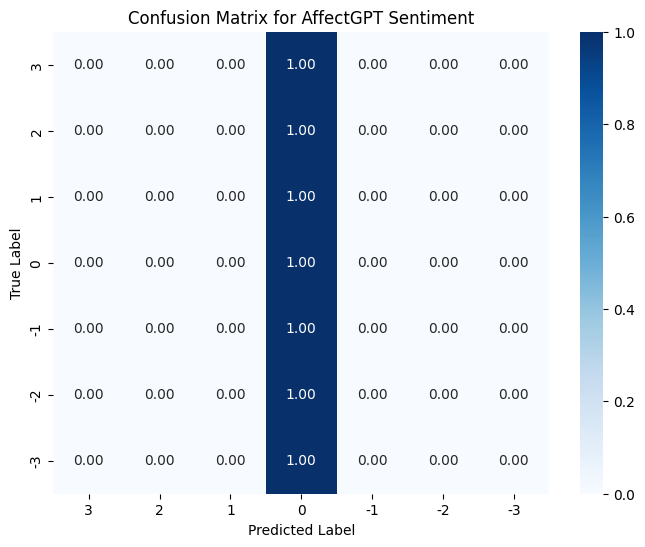}
        \caption{Confusion Matrix of AffectGPT (without caption) output on Sentiment Analysis Task (7-class).}
        \label{fig:conf-sent-aff}
    \end{minipage}\hfill
    \begin{minipage}[t]{.45\textwidth}
        \centering
        \includegraphics[width=1\linewidth]{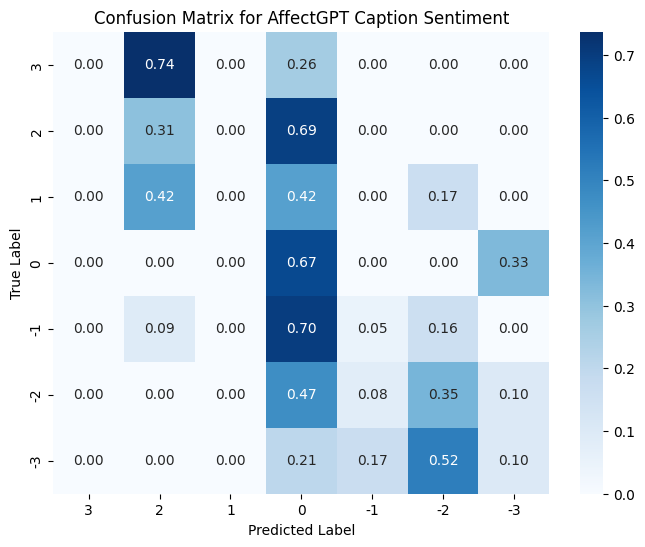}
        \caption{Confusion Matrix of AffectGPT with caption output on Sentiment Analysis Task (7-class).}
        \label{fig:conf-sent-aff-cap}
    \end{minipage}%

\end{figure}

\newpage

\begin{figure}[h]
    \centering
    \begin{minipage}[t]{.45\textwidth}
        \centering
        \includegraphics[width=1\linewidth]{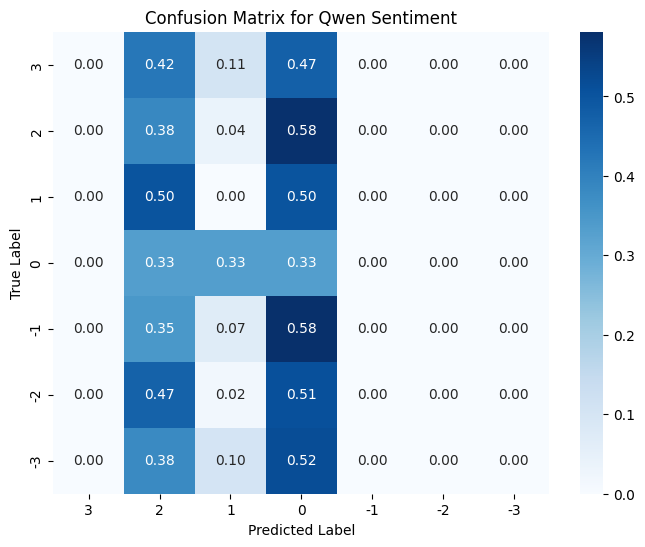}
        \caption{Confusion Matrix of Qwen-2.5 (without caption) output on Sentiment Analysis Task (7-class).}
        \label{fig:conf-sent-qwen}
    \end{minipage}\hfill
    \begin{minipage}[t]{.45\textwidth}
        \centering
        \includegraphics[width=1\linewidth]{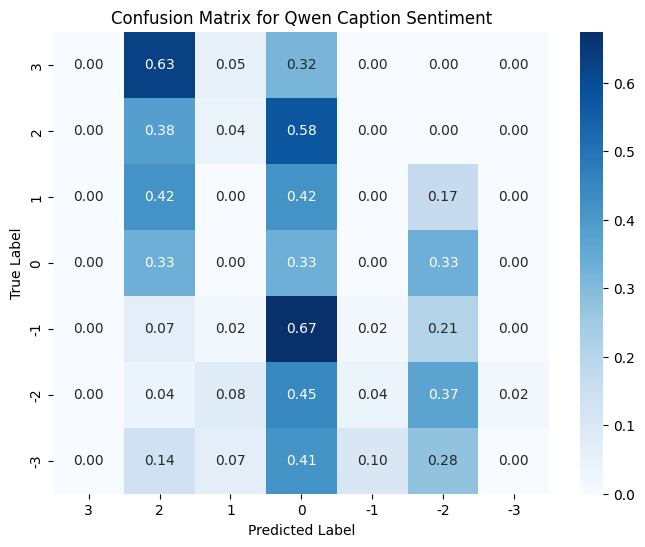}
        \caption{Confusion Matrix of Qwen-2.5 with caption output on Sentiment Analysis Task (7-class).}
        \label{fig:conf-sent-qwen-cap}
    \end{minipage}%

\end{figure}

\begin{figure}[h]
    \centering
    \begin{minipage}[t]{.45\textwidth}
        \centering
        \includegraphics[width=1\linewidth]{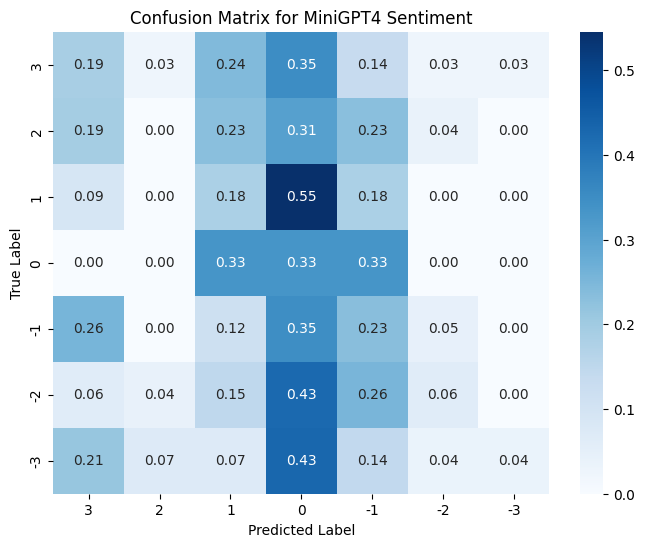}
        \caption{Confusion Matrix of MiniGPT4 (without caption) output on Sentiment Analysis Task (7-class).}
        \label{fig:conf-sent-mini}
    \end{minipage}\hfill
    \begin{minipage}[t]{.45\textwidth}
        \centering
        \includegraphics[width=1\linewidth]{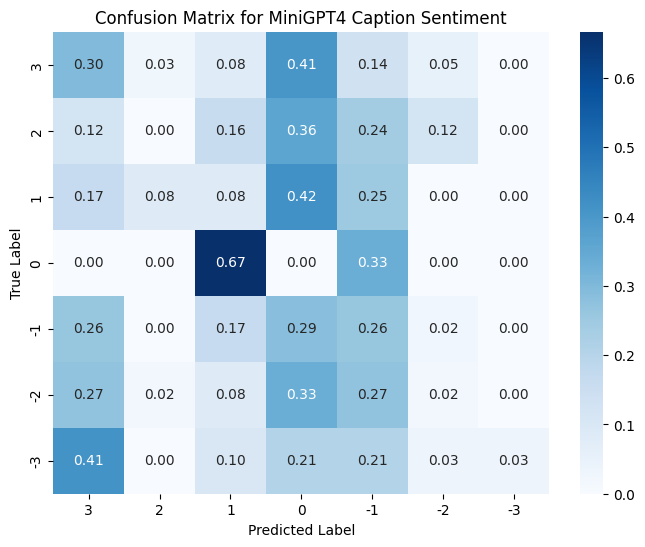}
        \caption{Confusion Matrix of MiniGPT4 with caption output on Sentiment Analysis Task (7-class).}
        \label{fig:conf-sent-mini-cap}
    \end{minipage}%

\end{figure}

\newpage

\begin{figure}[h]
    \centering
    \includegraphics[width=.8\linewidth]{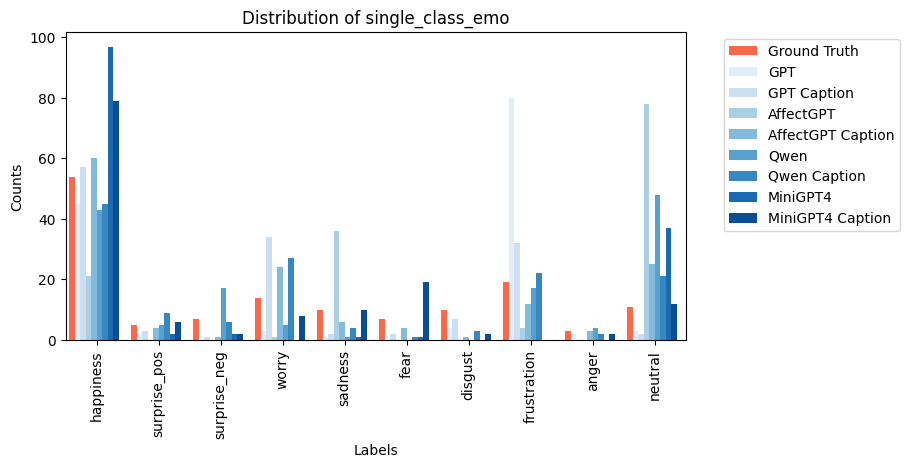}
    \caption{Distribution of Single Expression Emotion Classification Results.}
    \label{fig:emo-dist}
\end{figure}

\begin{figure}[h]
    \centering
    \begin{minipage}[t]{.45\textwidth}
        \centering
        \includegraphics[width=1\linewidth]{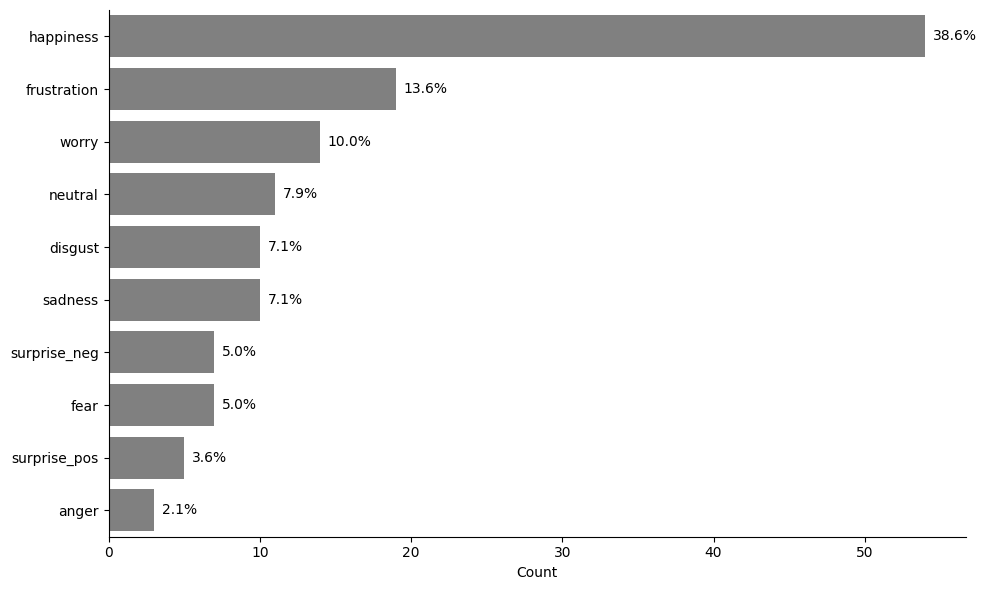}
        \caption{Distribution of emotion categories within the single expression set. Numbers above the bars indicate count.}
        \label{fig:single-exp-distrib}
    \end{minipage}\hfill
    \begin{minipage}[t]{.45\textwidth}
        \centering
        \includegraphics[width=1\linewidth]{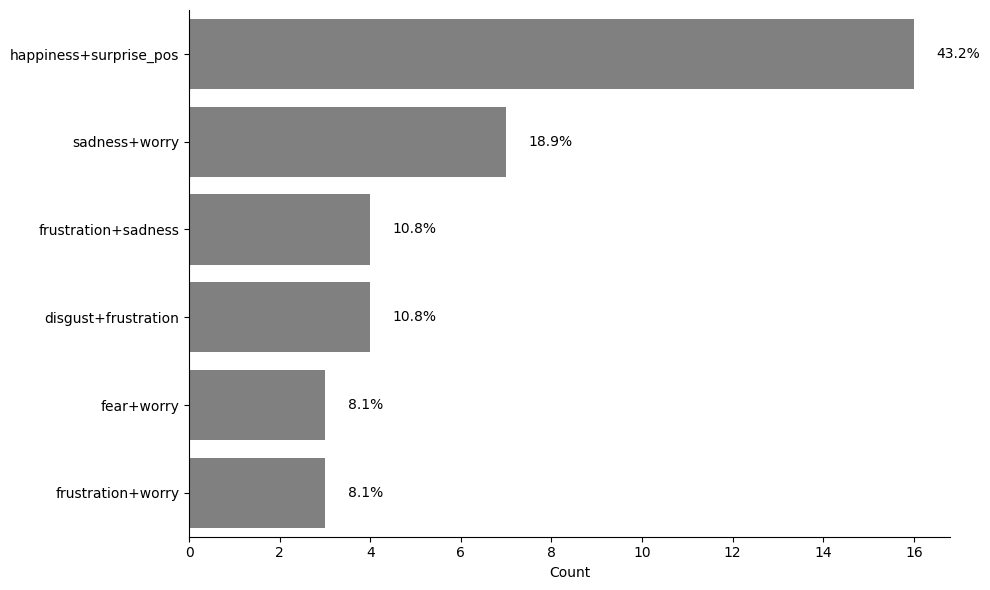}
        \caption{Distribution of emotion categories within the multi expression set. Numbers above the bars indicate count.}
        \label{fig:multi-exp-distrib.png}
    \end{minipage}%

\end{figure}

\begin{figure}[h]
    \centering
    \begin{minipage}[t]{.45\textwidth}
        \centering
        \includegraphics[width=1\linewidth]{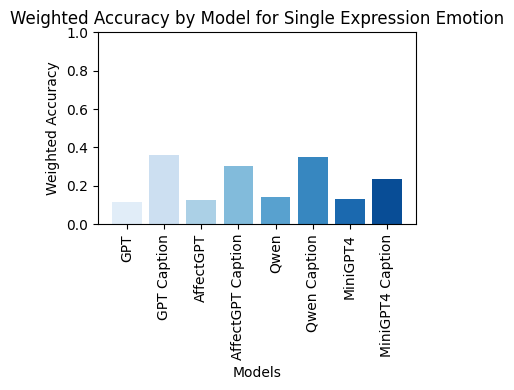}
        \caption{Weighted Accuracy of Models on Single Expression Emotion Classification Task.}
        \label{fig:emo-wacc}
    \end{minipage}\hfill
    \begin{minipage}[t]{.45\textwidth}
        \centering
        \includegraphics[width=1\linewidth]{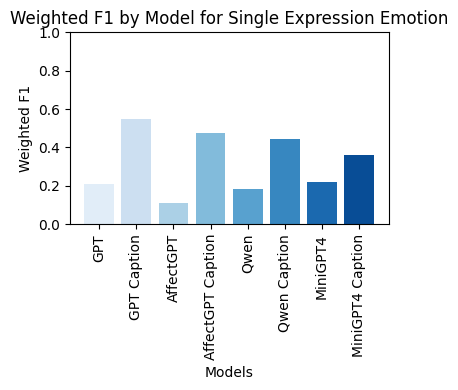}
        \caption{Weighted Accuracy of Models on Single Expression Emotion Classification Task.}
        \label{fig:emo-wf1}
    \end{minipage}%

\end{figure}

\newpage

\begin{figure}[h]
    \centering
    \begin{minipage}[t]{.45\textwidth}
        \centering
        \includegraphics[width=1\linewidth]{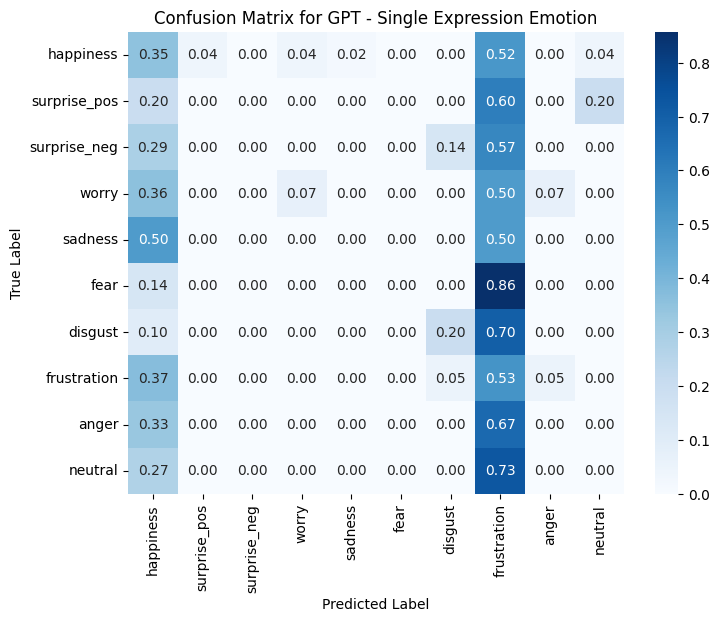}
        \caption{Confusion Matrix of GPT (without caption) output on Single Expression Emotion Classification Task.}
        \label{fig:conf-emo-gpt}
    \end{minipage}\hfill
    \begin{minipage}[t]{.45\textwidth}
        \centering
        \includegraphics[width=1\linewidth]{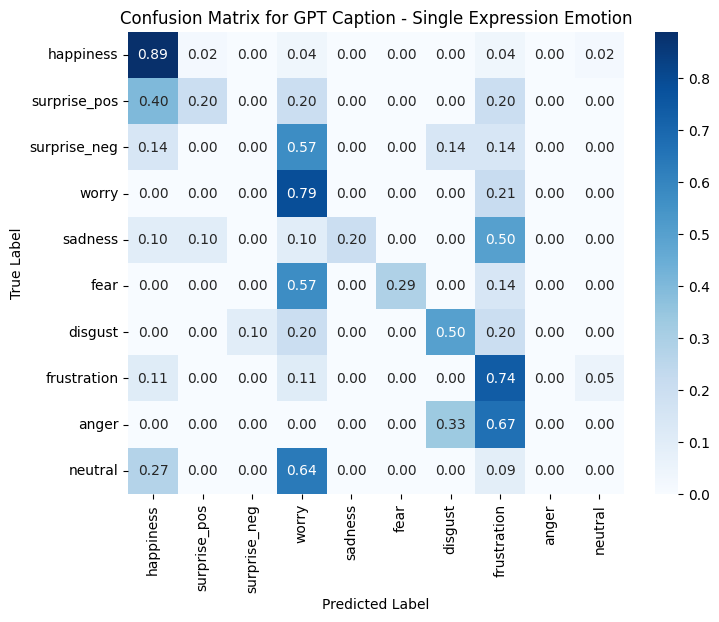}
        \caption{Confusion Matrix of GPT with caption output on Single Expression Emotion Classification Task.}
        \label{fig:conf-emo-gpt-cap}
    \end{minipage}%

\end{figure}

\begin{figure}[h]
    \centering
    \begin{minipage}[t]{.45\textwidth}
        \centering
        \includegraphics[width=1\linewidth]{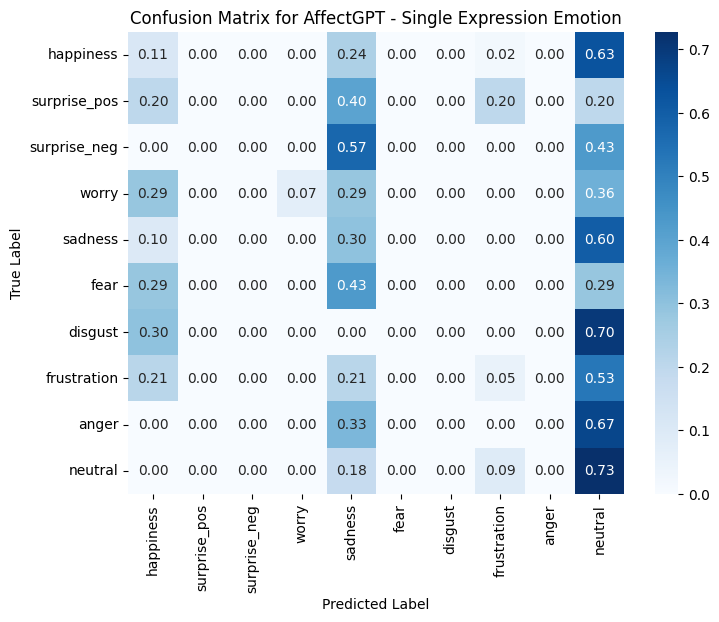}
        \caption{Confusion Matrix of AffectGPT (without caption) output on Single Expression Emotion Classification Task.}
        \label{fig:conf-emo-aff}
    \end{minipage}\hfill
    \begin{minipage}[t]{.45\textwidth}
        \centering
        \includegraphics[width=1\linewidth]{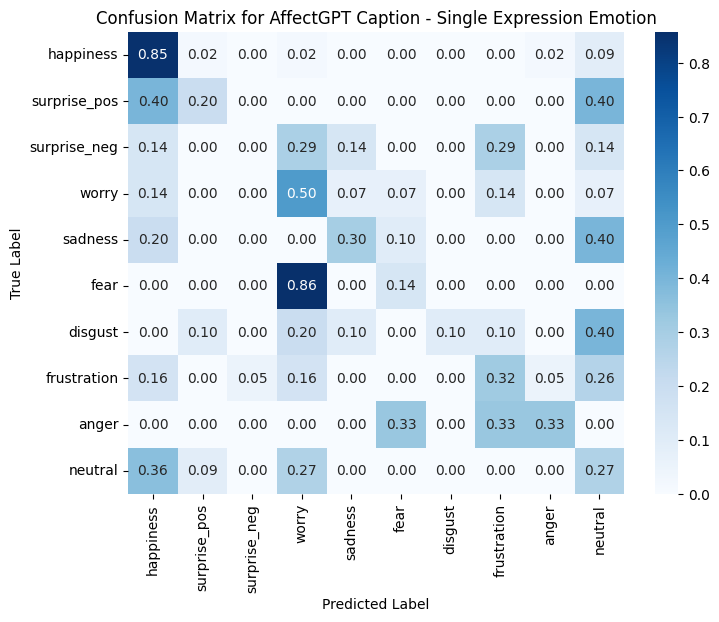}
        \caption{Confusion Matrix of AffectGPT with caption output on Single Expression Emotion Classification Task.}
        \label{fig:conf-emo-aff-cap}
    \end{minipage}%

\end{figure}

\newpage

\begin{figure}[h]
    \centering
    \begin{minipage}[t]{.45\textwidth}
        \centering
        \includegraphics[width=1\linewidth]{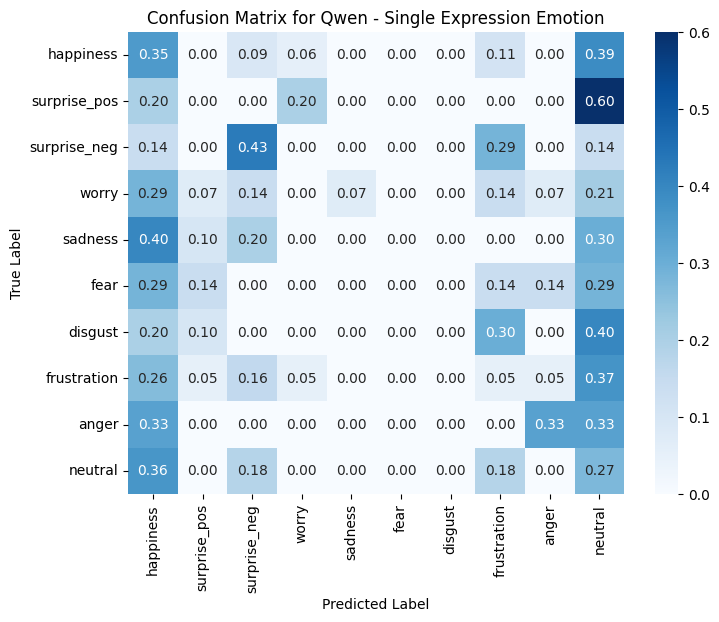}
        \caption{Confusion Matrix of Qwen-2.5 (without caption) output on Single Expression Emotion Classification Task.}
        \label{fig:conf-emo-qwen}
    \end{minipage}\hfill
    \begin{minipage}[t]{.45\textwidth}
        \centering
        \includegraphics[width=1\linewidth]{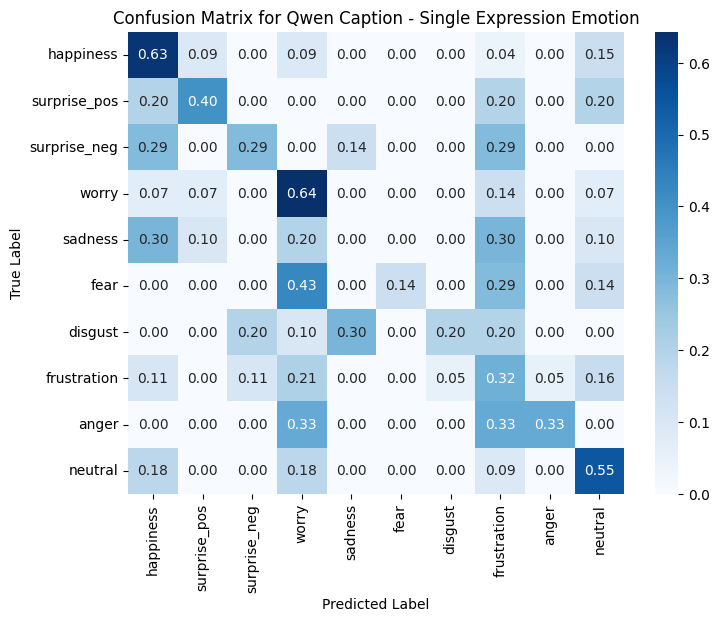}
        \caption{Confusion Matrix of Qwen-2.5 with caption output on Single Expression Emotion Classification Task.}
        \label{fig:conf-emo-qwen-cap}
    \end{minipage}%

\end{figure}

\begin{figure}[h]
    \centering
    \begin{minipage}[t]{.45\textwidth}
        \centering
        \includegraphics[width=1\linewidth]{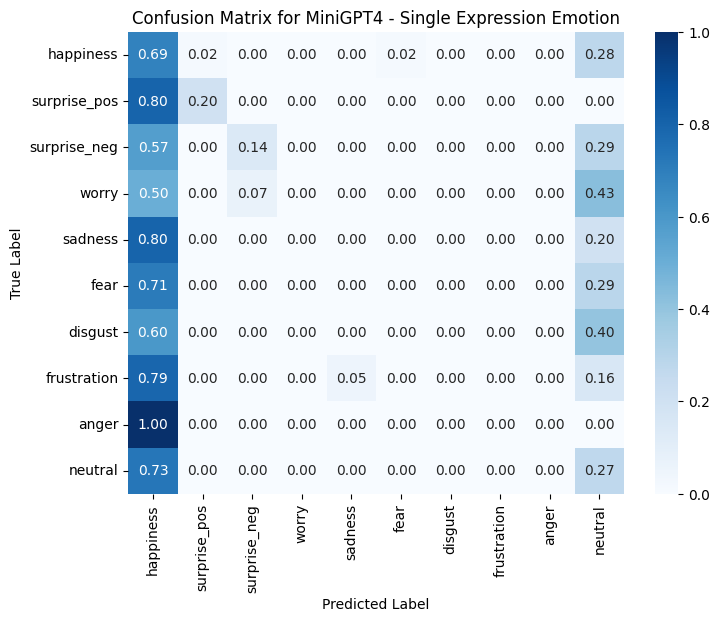}
        \caption{Confusion Matrix of MiniGPT4 (without caption) output on Single Expression Emotion Classification Task.}
        \label{fig:conf-emo-mini}
    \end{minipage}\hfill
    \begin{minipage}[t]{.45\textwidth}
        \centering
        \includegraphics[width=1\linewidth]{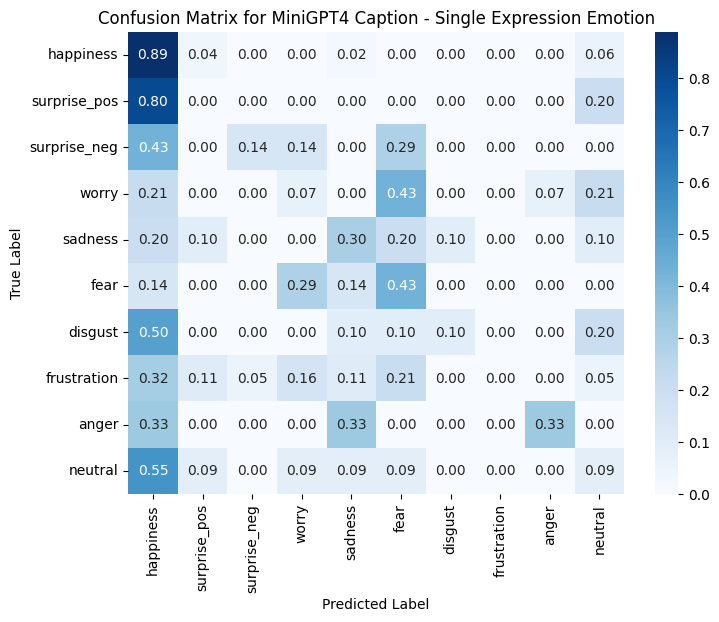}
        \caption{Confusion Matrix of MiniGPT4 with caption output on Single Expression Emotion Classification Task.}
        \label{fig:conf-emo-mini-cap}
    \end{minipage}%

\end{figure}

\end{document}